%% file: paper.tex
\documentclass[]{bytedance_seed}

\usepackage[toc,page,header]{appendix}

\usepackage{minitoc}
\usepackage{booktabs}
\usepackage{array}
\usepackage{multirow}
\usepackage{siunitx}
\usepackage{caption}
\usepackage{nicematrix}
\usepackage{amsmath}
\usepackage{amssymb}
\usepackage{graphicx}
\usepackage{subcaption}
\usepackage{mathrsfs}

\title{SciDA: Scientific Dynamic Assessor of LLMs}

\author[]{ByteDance Seed}
\author[]{Peking University}
\author[]{M-A-P}

\abstract{

Advancement in Large Language Models (LLMs)  reasoning capabilities enables them to solve scientific problems with enhanced efficacy. Thereby, a high-quality benchmark for comprehensive and appropriate assessment holds significance, while existing ones either confront the risk of data contamination or lack involved disciplines. To be specific, due to the data source overlap of LLMs training and static benchmark, the keys or number pattern of answers inadvertently memorized (i.e. data contamination), leading to systematic overestimation of their reasoning capabilities, especially numerical reasoning. 

We propose \textbf{SciDA}, a multidisciplinary benchmark that consists exclusively of over 1k Olympic-level numerical computation problems, allowing randomized numerical initializations for each inference round to avoid reliance on fixed numerical patterns. We conduct a series of experiments with both closed-source and open-source top-performing LLMs, and it is observed that the performance of LLMs drop significantly under random numerical initialization. Thus, we provide truthful and unbiased assessments of the numerical reasoning capabilities of LLMs. The data is available at \hyperlink{https://huggingface.co/datasets/m-a-p/SciDA}{https://huggingface.co/datasets/m-a-p/SciDA}

}

\date{\today}
\correspondence{ \\ Junting Zhou at \email{juntingzhou@stu.pku.edu.cn}, Ge Zhang at \email{zhangge.eli@bytedance.com}}

\begin{document}
\maketitle


\input{sections/introduction}

\input{sections/relatedwork}

\newpage
\input{sections/approach}
\input{sections/experiments}

\newpage
\input{sections/discussion}

\input{sections/conclusion}

\input{sections/future_work}

\clearpage

\bibliographystyle{plainnat}
\bibliography{main}

\clearpage

\beginappendix

\input{sections/appendix}

\end{document}

%% file: sections/introduction.tex

\section{Introduction}

Recent advancements in large language models (LLMs) have demonstrated remarkable capabilities in complex reasoning tasks. However, evaluating their true reasoning capabilities remains challenging, particularly in scientific domains, which require multi-step calculation and symbolic manipulation.

To assess the problem-solving capabilities of LLMs quantitatively, a series of benchmarks (GSM8k, MATH, MMLU, etc.) have been created and widely applied \citep{cobbe2021verifier,hendrycks2021math,hendrycks2021mmlu}, with data primarily sourced from academic competition questions and textbooks. These early-stage benchmarks have become relatively easy for frontier-level LLMs. Further works (GPQA, MMLU-Pro, Agieval, Scibench, Scieval, etc.) enable more comprehensive and rigorous assessment by incorporating more knowledge domains and diversifying data sources \citep{rein2024gpqa,wang2024mmlupro,zhong2023agieval,wang2023scibench,sun2024scieval}. However, there has been an obscured essential contradiction: open-access textbooks, examination questions, academic literature, and online datasets, which are the primary data sources of benchmarks, also serve as the data sources of LLMs' pretraining and fine-tuning. As a result, data leakage and contamination are highly probable, and certain combinations of numbers can be memorized, thereby hindering their ability to generalize or leading to a systematic overestimation of their cognitive reasoning capabilities \citep{golchin2023time, deng2023conta,dong2024generalization}. This is particularly concerning in domains requiring numerical reasoning, where reliance on memorized answers or combinations of numbers could have a more pronounced influence.

Generative benchmarks like KORgym\citep{shi2025korgymdynamicgameplatform} have been released recently. However, those works focus on game-based interactions, linguistic adaptability, or toy problems, lacking comprehensiveness and scientific rigor. In spite of mathematics, existing benchmarks lack assessment across various branches of natural science (physics, chemistry, biology, etc.), while such capabilities are crucial for real-world applications in scientific research.
This gap underscores the need for a comprehensive multi-disciplines dynamically generated benchmark to truthfully reflect the complexity and unpredictability of scientific problem-solving.

\begin{figure}[htbp]
    \centering
    \includegraphics[width=\textwidth]{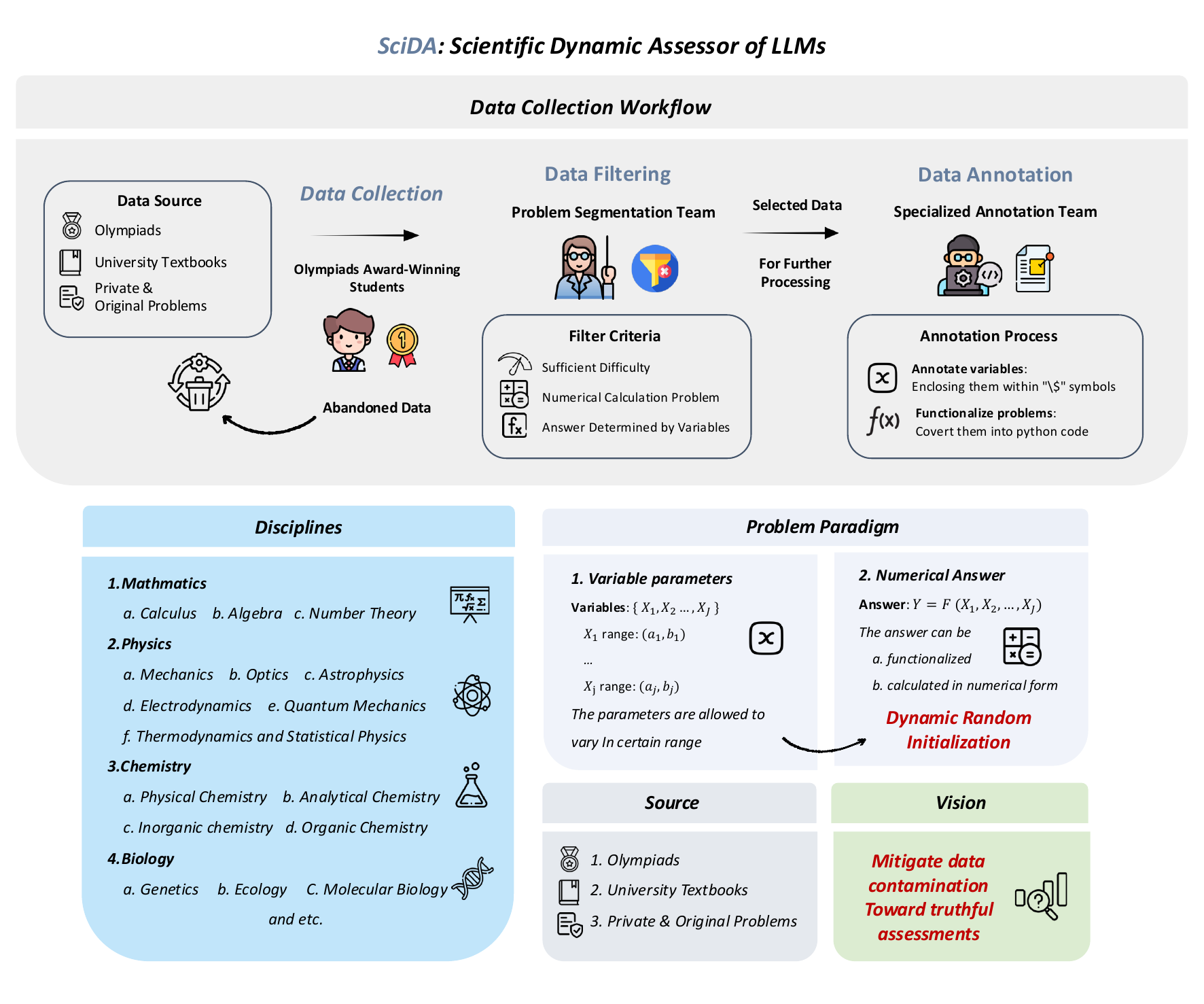} 
    \caption{The data construction pipeline of the SciDA. \textit{\textbf{Data Collection Workflow}} illustrates the process of collecting and filtering scientific problems from Olympiad competitions and university textbooks, followed by variable annotation and numerical functionalization. \textit{\textbf{Discipline}} shows that our benchmark covers various subjects include mathematics, physics, chemistry, and biology. \textit{\textbf{Problem Paradigm}} shows that the benchmark supports dynamic random initialization and aims to provide a robust and contamination-free evaluation for scientific reasoning models.}
    \label{fig:SciDA-main-fig}
\end{figure}

To address these limitations, we introduce \textbf{SciDA}, a dynamic scientific benchmark built on 1,000 expert-curated problems from Olympiad-level competitions spanning mathematics, physics, chemistry, and biology. Each problem undergoes structured variable extraction, where all modifiable parameters are programmatically identified and replaced with \$ tokens (e.g., \$m\$, \$k\$, \$$v_0$\$). During each evaluation iteration, these tokens are dynamically initialized with randomized values sampled from predefined scientifically valid ranges. Our data collection pipeline prioritizes quality, diversity, and complexity through domain-expert annotation, symbolic consistency verification, range validation for randomized variables, and solvability checks across value permutations.

To our best knowledge, SciDA is the first dynamic, contamination-proof benchmark for rigorous scientific reasoning evaluation. Our work yields three pivotal insights:

\begin{itemize}
  \item Data leakage is a widespread issue in large language models, raising concerns about fairness and the validity of model evaluation.
  
  \item The generalization ability of large language models varies across disciplines when parameters are randomly initialized, indicating that domain-specific factors significantly influence model performance.
  
  \item The use of code interpreters substantially impacts the robustness of model-based computation, suggesting the necessity of integrating external tools to ensure accuracy and reliability in reasoning tasks.
\end{itemize}
\textbf{}

%% file: sections/relatedwork.tex
\section{Related Work}
\subsection{Scientific Benchmarks}
To comprehensively evaluate the performance of current LLMs, series of benchmarks (GSM8k, MATH, MMLU, etc.) have been created \citep{cobbe2021verifier,hendrycks2021math,hendrycks2021mmlu}, with data primarily sourced from academic competition questions and textbooks. Further works (GPQA, SuperGPQA, MMLU-Pro, Agieval, Scibench, Scieval, etc.) enable more comprehensive and rigorous assessment by incorporating more disciplines and diversifying data sources \citep{rein2024gpqa,pteam2025supergpqascalingllmevaluation, wang2024mmlupro,zhong2023agieval,wang2023scibench,sun2024scieval}. However, owing to the advancement in LLMs capabilities, existing benchmarks have become relatively easy for advanced LLMs and existing benchmarks. Therefore, the need for scientific benchmarks to assess the limits of advanced LLMs naturally emerged.

Driven by such need, a major focus is to collect problems with higher complexity, primarily Olympics problems, i.e. "the pearl of human wisdom". Undergraduate-level \citep{liu2024mathbench,tang2024mathscale} and Olympic-level \citep{he2024olympiadbench, huang2025olympicarena, sun2025olymmath} benchmarks are created and applied. For instance, OlymMATH\citep{sun2025olymmath} is a benchmark Olympics-level mathematical problems spanning multiple mathematical domains, including algebra and geometry. Omni-MATH\citep{gao2024omnimath} is also a mathematical benchmark at the Olympic level integrated with a data leakage detection mechanism, on which the most advanced LLMs (such as OpenAI's o1) achieve accuracy rates of only 52.55\%. 

Another focus is to expand the scope of disciplines and go beyond mathematics. Both multidisciplinary \citep{wang2024mmlupro,wang2023scibench,sun2024scieval,huang2025olympicarena} and discipline specific benchmarks\citep{jain2025livecodebench,qiu2025phybench} emerge. For instance, SciEval\citealp{sun2024scieval} includes disciplines of physics, chemistry and biology and OlympicArena \citep{huang2025olympicarena} spans 7 core disciplines: mathematics, physics, chemistry, biology, geography, astronomy, and computer science. Meanwhile, some benchmarks feature domain specific and focus on relatively naive disciplines, such as PHYbench \citep{qiu2025phybench} that consists of 500 original physics problems, which fills the blank of high-quality physics benchmarks.

\subsection{Dynamic Benchmark}
To mitigate data contamination, some researchers update parameters manually and periodically. That is, to create dynamic benchmarks, such as VarBench \citep{qian2024varbench} and LiveCodeBench \citep{jain2025livecodebench}), of which the parameters are variable rather than constant. However, such dynamism is artificially maintained pseudo-dynamism. Live updates come with the burden of sustained collection and processing of high-quality data. Thus, the solution is expedient, evading the essential issue.

Further works like KORgym\citep{shi2025korgymdynamicgameplatform} do realize dynamic initialization, but they primarily focus on game-based interactions, linguistic adaptability, or toy problems, lacking rigor and not being able to accurately assess the capabilities of scientific problem-solving. Meanwhile, benchmarks like Math-perturb \citep{huang2025mathperturbbenchmarkingllmsmath}, which concerns purely mathematics, lack comprehensiveness and are limited to few discplines. Comprehensive assessment across various branches of natural science (physics, chemistry, biology, etc.) is vital, since such capabilities are crucial for real-world applications in scientific research.

Despite the effectiveness in mitigating data contamination, existing dynamics benchmarks are inconsistent and unsatisfactory in form and quality, while remaining limited to few disciplines, primarily mathematics.


%% file: sections/approach.tex
\section{Approach}

\subsection{Problem Formalization}

Let \(q \sim \mathcal{Q}\) denote a problem sampled from the problem distribution \(\mathcal{Q}\).

Suppose \(Q\) contains \(J\) random variables, indexed by \(i = 1,2,\ldots,J\). Denote these random variables by
\[
\{\,\mathcal{X}_1, \mathcal{X}_2, \ldots, \mathcal{X}_J\,\}.
\]
Each \(\mathcal{X}_i\) is drawn from a uniform distribution over the interval \([\,a,b\,]\):
\[
\mathcal{X}_i \sim \mathcal{U}(a,b),
\quad
\]
where \(a\) is the minimum possible value and \(b\) is the maximum possible value, determined by the actual meaning of each variable.
\medskip

\textbf{Initialization.}
When initializing the problem, each random variable \(\mathcal{X}_i\) is independently sampled to obtain a realization \(x_i\):
\[
x_i \sim \mathcal{U}(a,b),
\quad
\forall\,i \in \{1,2,\ldots,J\}.
\]

Therefore, all variables in problem $q$ are randomly initialized before reasoning.

\textbf{Answer generation.}
After the initialization of \(\{x_1, x_2, \ldots, x_J\}\), the correct answer $\mathbf{y}$ to problem \(Q\) is designed to be a finite sequence of real numbers.  Formally, there exists a known, labeled function
\[
F : \mathbb{R}^J \;\longrightarrow\; \mathbb{R}^K,
\]
such that
\[
\mathbf{y} \;=\; F\bigl(x_1,\,x_2,\,\ldots,\,x_J\bigr)
\]
is the ground‐truth answer vector of length \(K\).  Here,
\[
\mathbf{y} \;=\; \bigl(y_1,\,y_2,\,\ldots,\,y_K\bigr) \in \mathbb{R}^K
\]
depends deterministically on the initialized values \((x_1, x_2, \ldots, x_J)\).

\medskip

\textbf{Model Prediction and Correctness Criterion.}  
Let \(\widehat{\mathbf{y}} = \bigl(\hat y_1,\,\hat y_2,\,\ldots,\,\hat y_K\bigr)\) denote the sequence of numbers predicted by the model (or solver) in response to \(Q\).  We say that the model’s answer is \emph{correct} if its deviation from the true answer \(\mathbf{y}\) is within a prescribed tolerance.  
\begin{figure}[h!tbp]
    \centering
    \begin{subfigure}[b]{0.45\textwidth}
        \includegraphics[width=\textwidth]{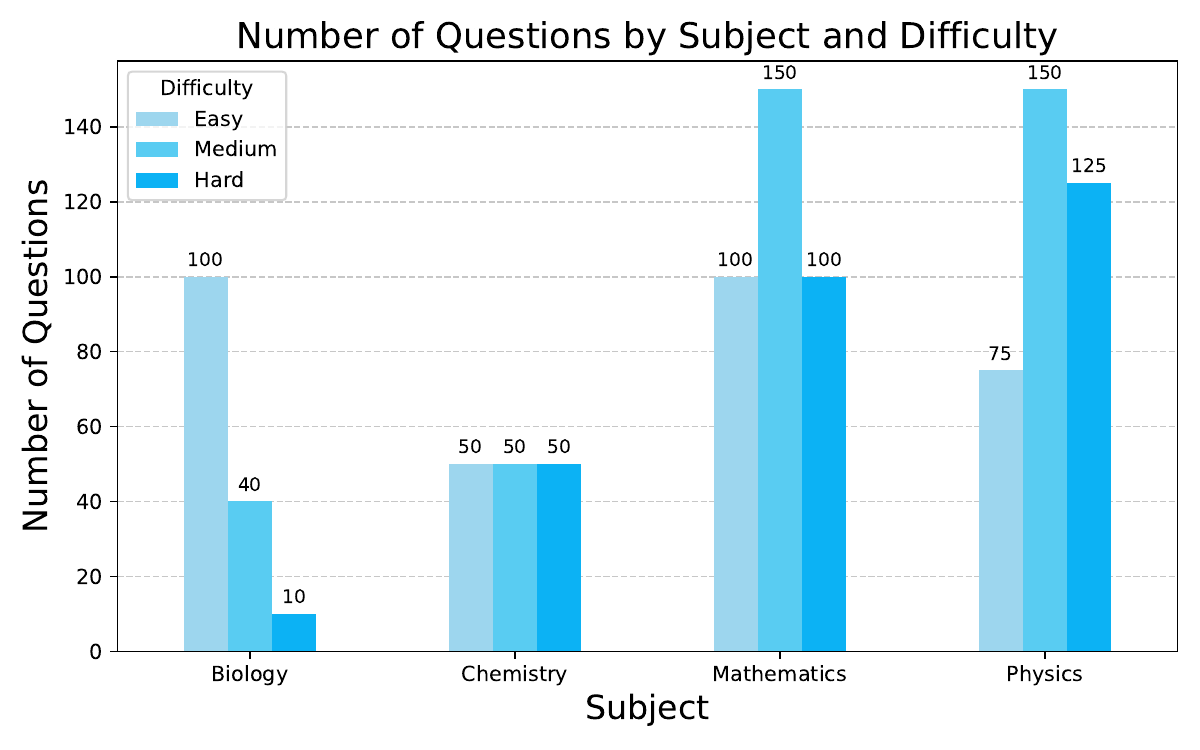}
        \caption{Disciplinary and difficulty distribution of the collected dataset.}
        \label{fig:data_distribution}
    \end{subfigure}
    \hfill
    \begin{subfigure}[b]{0.5\textwidth}
        \includegraphics[width=\textwidth]{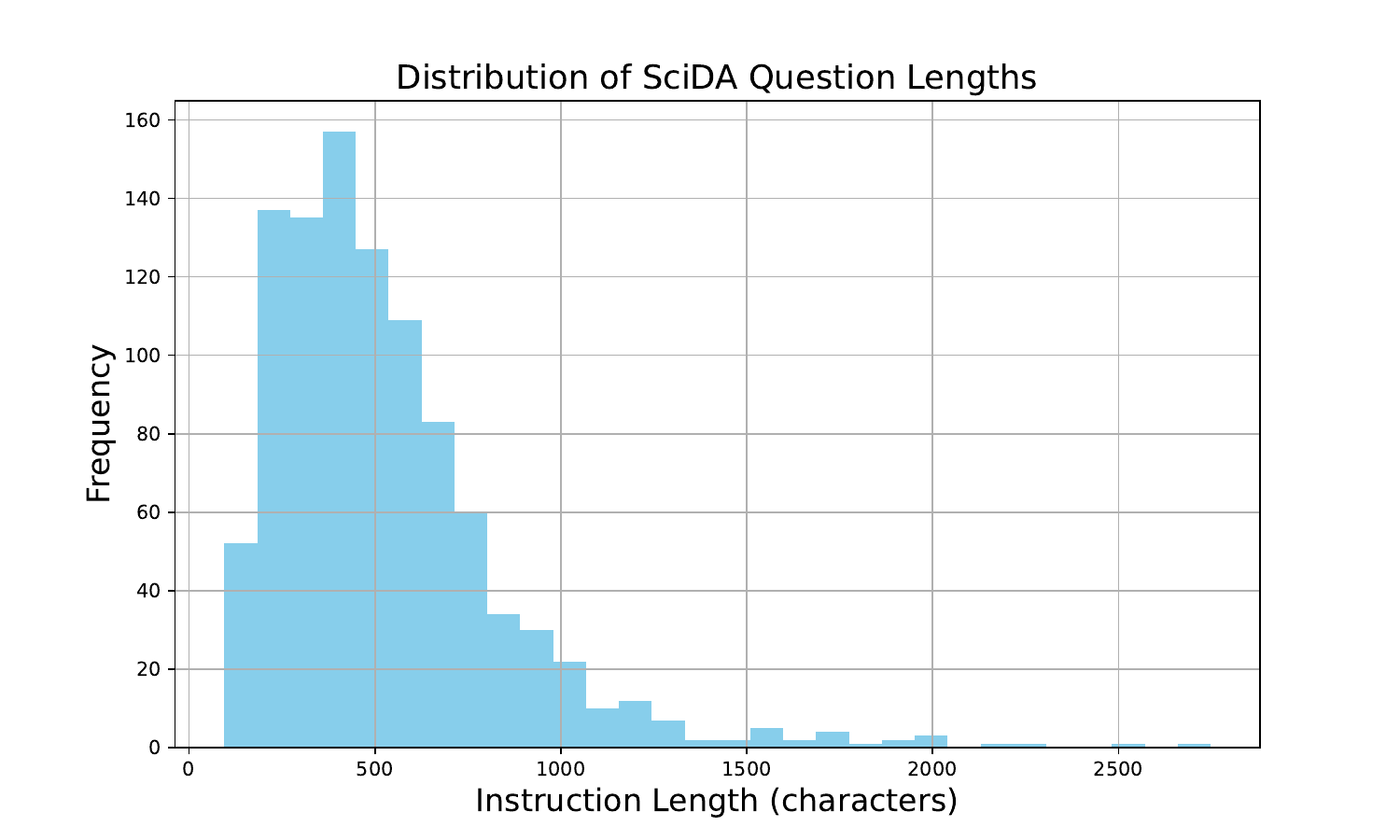}
        \caption{Detailed difficulty information for the problems in the dataset.}
        \label{fig:length_info}
    \end{subfigure}
    \caption{Distribution and difficulty information of the collected dataset.}
    \label{fig:combined_distribution}
\end{figure}

\subsection{Data Collection}

Our data collection process involved three main steps to create a high-quality dataset of variable-based computational problems.

First, a team of students with competition backgrounds meticulously collected problems from regional and international Olympiad competition problems, Olympiad workbook and guides, related online platforms and university textbooks. This broad collection ensured a diverse initial pool of potential problems.

Next, a problem-segmentation team filtered this initial pool based on three strict criteria: 1) sufficient difficulty, 2) being a computational problem containing variables, and 3) the answer being determined by variables and presented in a numerical format. Problems satisfying these criteria were extracted for the next stage.

\begin{figure}[htbp]
    \centering
    \includegraphics[width=1.0\textwidth]{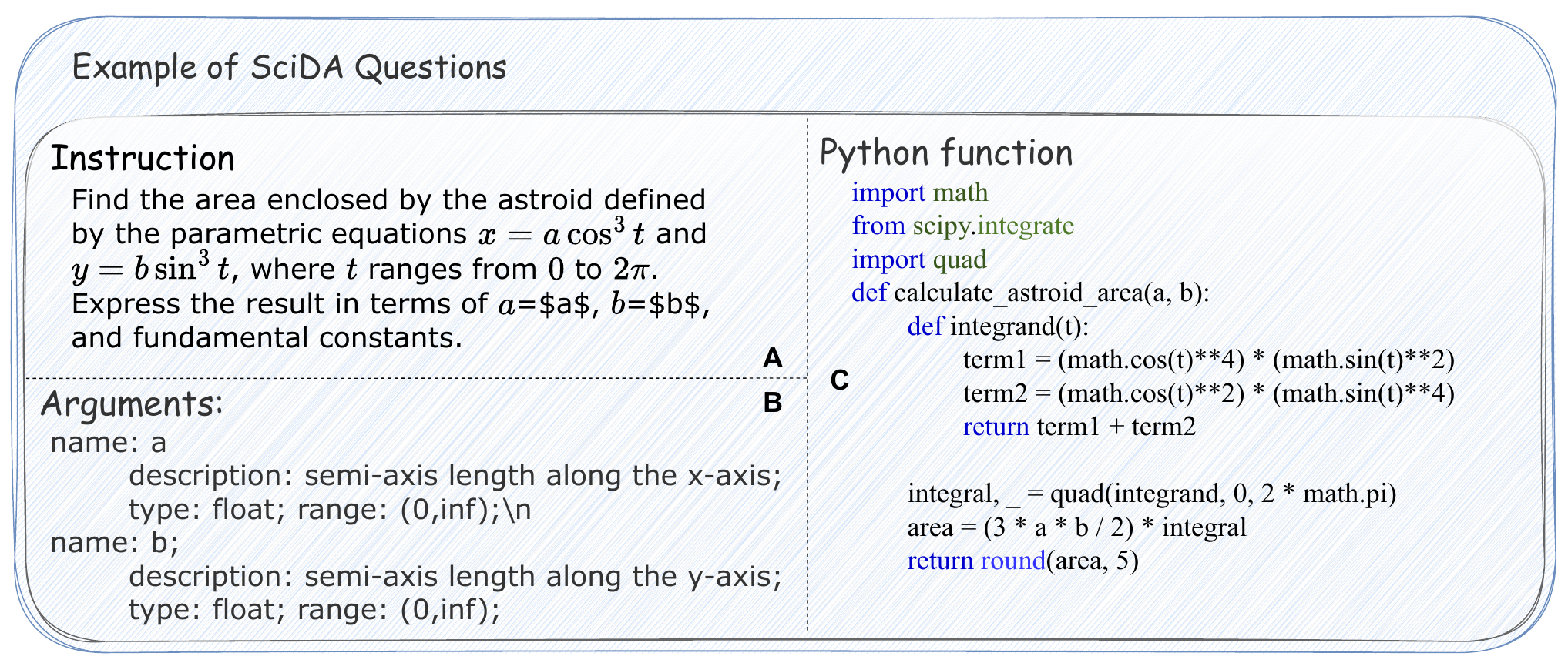} 
    \caption{Example of a \textbf{SciDA} problem. (A) shows the problem instruction with variables annotated using "\$" symbols. (B) shows how the arguments are labeled. (C) shows the Python code to generate the answer.}
    \label{fig:example_problem}
\end{figure}

Finally, a specialized annotation team took the extracted problems and performed two key tasks: annotating the variables by enclosing them within "\$" symbols (as exemplified in Figure \ref{fig:example_problem}, and writing Python code to solve each problem.

After completing the annotation, we subjected the variables in our problems to five rounds of initialization. This involved assigning different values to the variables and verifying that the corresponding Python code executed correctly and produced valid results. Following this rigorous cleaning process, we obtained a dataset of 1000 problems that met all our specified criteria. The disciplinary and difficulty distributions of this dataset are presented in Figure \ref{fig:data_distribution}. Statistics on the length of the questions are shown in Figure \ref{fig:length_info}. 




%% file: sections/experiments.tex
\section{Experiments}
We select initialization parameters and random parameters 5 times (the reason why we chose this hyperparameter is described in \ref{appendix:Elbow_Analysis}), and selected 14 mainstream models to conduct the experiments. Model performance is shown in Table \ref{tab:model-performance}.


\textbf{Overall Comparison of Models.}

Generally, the accuracy of various models on our benchmark with initial parameters ranges from 20\% to 50\%, which demonstrates that our benchmark is sufficiently challenging and the problems selected are of satisfactory quality. Under random initialization, the accuracy drop of those models ranges from 10\% to 20\%, which indicates a significant decrease of 20\% to 60\% relatively.

The performance of different models on the benchmark exhibits significant variation. Gemini-2.5-pro OpenAI-o3 and Doubao1.5-pro-thinking are the best-performing ones, with initial accuracy rates of approximately 50\% and randomized accuracy rates of around 35\%. In contrast, the older model GPT4o has the lowest accuracy, with initial and randomized accuracy rates both hovering around 25\% and 10\%, respectively. 

To conclude, our scientific benchmark is satisfactory in difficulty and discriminability, thus can be used to scientifically and comprehensively assess the capabilities of reasoning and computation of models.

\textbf{Thinking models perform better in calculation problems that Instruct models.}

When dealing with our selected long-chain reasoning and calculation problems, Thinking models outperform Instruct models. More specifically, Gemini-2.5-pro, OpenAI-o3 and Doubao1.5-pro-thinking have superior performance. We attribute this to their enhanced capacity for slow thinking, exemplified by the application of chain-of-thought reasoning, multi-step inference, and emphasis on logical coherence. This observation underscores the necessity of slow thinking and, by extension, highlights the importance of utilizing high-quality, curated, and logically robust training data during the training process, which facilitates models' acquisition of "slow thinking" strategies.

We are convinced that applying this benchmark to reinforcement learning holds considerable potential in promoting LLMs to engage in slow thinking and enhancing their reasoning capabilities.

\begin{table}[htbp]

\centering
\caption{Model Performance on SciDA}
\label{tab:model-performance}
\resizebox{\textwidth}{!}{%
\begin{tabular}{@{}l *{9}{S[table-format=2.2]}@{}}
\toprule
\textbf{Model Name} &
\multicolumn{1}{c}{\textbf{Avg}} &
\multicolumn{4}{c}{\textbf{Initial}} &
\multicolumn{4}{c}{\textbf{Random}} \\
\cmidrule(lr){3-6} \cmidrule(lr){7-10}
& &
\multicolumn{1}{c}{\textbf{All}} &
\multicolumn{1}{c}{\textbf{Easy}} &
\multicolumn{1}{c}{\textbf{Medium}} &
\multicolumn{1}{c}{\textbf{Hard}} &
\multicolumn{1}{c}{\textbf{All}} &
\multicolumn{1}{c}{\textbf{Easy}} &
\multicolumn{1}{c}{\textbf{Medium}} &
\multicolumn{1}{c}{\textbf{Hard}} \\
\midrule
Gemini-2.5-pro.preview.0506.google.ci & \textbf{40.19} & 49.84 & 55.68 & 48.68 & \textbf{42.54} & \textbf{38.26} & 42.05 & \textbf{38.04} & \textbf{32.63} \\

OpenAI-o3-high.code & 39.45 & \textbf{52.22} & 62.60 & \textbf{51.06} & 37.72 & 36.90 & 45.10 & 33.23 & 30.00 \\
Doubao1.5-pro-thinking.0415 & 37.64 & 51.50 & \textbf{64.27} & 47.88 & 37.28 & 34.87 & \textbf{45.43} & 30.32 & 25.70 \\
DeepSeek-reasoner-R1.volc & 35.37 & 47.98 & 59.28 & 44.18 & 36.40 & 32.84 & 41.44 & 28.78 & 25.96 \\
OpenAI-o4-mini.high.0416.code & 33.18 & 44.47 & 55.12 & 40.74 & 33.77 & 30.92 & 39.89 & 25.93 & 25.00 \\
Claude-gcp.37.thinking & 32.09 & 48.29 & 63.43 & 44.18 & 31.14 & 28.85 & 39.45 & 24.13 & 19.91 \\
Gemini-2.5-pro.preview.0506 & 30.20 & 46.12 & 53.19 & 46.30 & 34.65 & 27.01 & 32.08 & 24.66 & 22.89 \\
OpenAI-o1-1217.high.code & 29.21 & 41.78 & 51.25 & 39.95 & 29.82 & 26.70 & 34.46 & 22.49 & 21.40 \\
DeepSeek-V3-0324.volc.forCompetitor & 26.65 & 41.99 & 54.85 & 38.10 & 28.07 & 23.58 & 33.46 & 18.73 & 15.96 \\
OpenAI-o3-mini.high.code & 25.58 & 42.30 & 52.91 & 37.83 & 32.89 & 22.23 & 27.98 & 18.20 & 19.82 \\
Gemini-2.5-flash.preview.0520 & 24.20 & 41.26 & 51.25 & 38.62 & 29.82 & 20.79 & 27.31 & 17.41 & 16.05 \\
GPT4o-1120 & 14.82 & 28.54 & 40.99 & 25.40 & 14.04 & 12.08 & 18.39 & 9.74 & 5.96 \\
GPT4o-0513 & 14.68 & 27.51 & 40.72 & 23.28 & 13.60 & 12.12 & 19.11 & 10.00 & 4.56 \\
GPT4o-0806 & 12.96 & 25.75 & 40.17 & 20.11 & 12.28 & 10.40 & 16.84 & 7.88 & 4.39 \\
\bottomrule
\end{tabular}%
}
\smallskip
\parbox{\textwidth}{\small Note: Scores represent performance percentages on the \textbf{SciDA} benchmark. "Initial" denotes fixed-parameter problems, "Random" denotes dynamically parameterized problems.}
\end{table}



\textbf{LLMs have a biased performance across different subjects.} 

In terms of the average accuracy, under random initialization, mathematics exhibits the most significant decline, followed by physics, while chemistry and biology are relatively less affected. More specifically, under random initialization, the accuracy rates of mathematics and physics decrease by 30\% to 70\% compared to the initial conditions, while the maximum decrease for biology and chemistry does not exceed 50\%.

We think the observed phenomenon is due to the fact that mathematics and physics problems often require longer chains of thought (CoT) and involve more variables. Meanwhile, there is a relatively homogeneous set of classic numerical patterns, which can be memorized and lead to inflated performance. Therefore, when these numerical patterns are disrupted, the true challenge of the problems is revealed, leading to a more obvious deviation in accuracy.

\begin{table}[h!]
    \centering
    \caption{Different Model Performance Data Summary by Model, Subject, Type, and Difficulty}
    \label{tab:api_performance_combined}
    \resizebox{0.95\textwidth}{!}{%
    \begin{tabular}{ll c cccc cccc}
        \toprule
        \textbf{Model Name} & \textbf{Subject} & \textbf{Avg} & \multicolumn{4}{c}{\textbf{Initial}} & \multicolumn{4}{c}{\textbf{Random}} \\
        \cmidrule(lr){4-7} \cmidrule(lr){8-11}
        & & & \textbf{All} & \textbf{Easy} & \textbf{Med} & \textbf{Hard} & \textbf{All} & \textbf{Easy} & \textbf{Med} & \textbf{Hard} \\
        \midrule
        OpenAI-o3-high.code & Biology & $55.10$ & $82.31$ & $82.47$ & $85.00$ & $70.00$ & $49.66$ & $53.20$ & $42.00$ & $46.00$ \\
        & Chemistry & $35.37$ & $48.98$ & $61.22$ & $46.94$ & $38.78$ & $32.65$ & $35.51$ & $36.33$ & $26.12$ \\
        & Physics & $29.84$ & $37.43$ & $48.74$ & $34.97$ & $23.61$ & $28.32$ & $37.14$ & $25.17$ & $20.00$ \\
        & Math & $43.90$ & $55.16$ & $60.42$ & $58.90$ & $44.33$ & $41.65$ & $51.67$ & $37.67$ & $37.73$ \\
        \midrule
        Doubao1.5-pro-thinking.0415 & Biology & $56.35$ & $82.31$ & $83.51$ & $87.50$ & $50.00$ & $51.16$ & $56.29$ & $41.00$ & $42.00$ \\
        & Chemistry & $35.26$ & $48.98$ & $55.10$ & $48.98$ & $42.86$ & $32.52$ & $38.37$ & $33.47$ & $25.71$ \\
        & Physics & $29.44$ & $38.92$ & $54.62$ & $33.57$ & $23.61$ & $27.54$ & $39.33$ & $22.24$ & $18.61$ \\
        & Math & $38.64$ & $51.62$ & $61.46$ & $50.68$ & $43.30$ & $36.05$ & $45.63$ & $34.25$ & $29.28$ \\
        \midrule
        DeepSeek-reasoner-R1.volc & Biology & $53.17$ & $74.83$ & $72.16$ & $85.00$ & $60.00$ & $48.84$ & $51.75$ & $41.50$ & $50.00$ \\
        & Chemistry & $34.58$ & $41.50$ &$53.06$ & $34.69$ & $36.73$ & $33.20$ & $35.10$ & $34.69$ & $29.80$ \\
        & Physics & $28.24$ & $37.13$ & $50.42$ & $34.27$ & $20.83$ & $26.47$ & $38.15$ & $21.40$ & $17.22$ \\
        & Math & $35.00$ & $49.85$ & $60.42$ & $45.89$ & $45.36$ & $32.04$ & $38.33$ & $30.55$ & $28.04$ \\
        \midrule
        OpenAI-o4-mini.high.0416.code & Biology & $49.55$ & $74.15$ & $75.26$ & $77.50$ & $50.00$ & $44.63$ & $48.45$ & $39.00$ & $30.00$ \\
        & Chemistry & $24.60$ & $31.97$ & $34.69$ & $28.57$ & $32.65$ & $23.13$ & $27.76$ & $21.63$ & $20.00$ \\
        & Physics & $23.60$ & $32.04$ & $43.70$ & $27.97$ & $20.83$ & $21.92$ & $31.09$ & $17.06$ & $16.39$ \\
        & Math & $39.23$ & $49.26$ & $59.38$ & $47.26$ & $42.27$ & $37.23$ & $48.33$ & $32.47$ & $33.40$ \\
        \midrule
        Claude-gcp.37.thinking & Biology & $57.14$ & $78.23$ & $81.44$ & $80.00$ & $40.00$ & $52.93$ & $56.91$ & $47.50$ & $36.00$ \\
        & Chemistry & $32.88$ & $47.62$ & $63.27$ & $48.98$ & $30.61$ & $29.93$ & $36.33$ & $29.39$ & $24.08$ \\
        & Physics & $27.15$ & $39.82$ & $50.42$ & $35.66$ & $30.56$ & $24.61$ & $35.46$ & $19.30$ & $17.22$ \\
        & Math & $25.76$ & $43.95$ & $61.46$ & $41.10$ & $30.93$ & $22.12$ & $28.33$ & $20.68$ & $18.14$ \\
        \midrule
        Gemini-2.5-pro.preview.0506 & Biology & $46.94$ & $70.75$ & $68.04$ & $82.50$ & $50.00$ & $42.18$ & $45.15$ & $36.00$ & $38.00$ \\
        & Chemistry & $24.94$ & $38.78$ & $34.69$ & $42.86$ & $38.78$ & $22.18$ & $25.31$ & $22.04$ & $19.18$ \\
        & Physics & $24.30$ & $38.02$ & $45.38$ & $38.46$ & $25.00$ & $21.56$ & $26.55$ & $18.04$ & $20.28$ \\
        & Math & $31.02$ & $46.61$ & $57.29$ & $45.21$ & $38.14$ & $27.91$ & $29.17$ & $28.90$ & $25.15$ \\
        \midrule
        OpenAI-o1-1217.high.code & Biology & $46.49$ & $77.55$ & $79.38$ & $77.50$ & $60.00$ & $40.27$ & $45.57$ & $29.50$ & $32.00$ \\
        & Chemistry & $19.84$ & $25.17$ & $28.57$ & $22.45$ & $24.49$ & $18.78$ & $20.41$ & $18.37$ & $17.55$ \\
        & Physics & $19.61$ & $27.54$ & $35.29$ & $25.17$ & $19.44$ & $18.02$ & $24.71$ & $15.10$ & $12.78$ \\
        & Math & $35.25$ & $47.49$ & $54.17$ & $50.00$ & $37.11$ & $32.80$ & $42.50$ & $29.18$ & $28.66$ \\
        \midrule
        DeepSeek-V3-0324.volc.forCompetitor & Biology & $51.36$ & $74.15$ & $75.26$ & $77.50$ & $50.00$ & $46.80$ & $50.10$ & $39.50$ & $44.00$ \\
        & Chemistry & $27.55$ & $41.50$ & $46.94$ & $40.82$ & $36.73$ & $24.76$ & $33.47$ & $22.04$ & $18.78$ \\
        & Physics & $19.81$ & $32.93$ & $45.38$ & $30.77$ & $16.67$ & $17.19$ & $26.55$ & $13.43$ & $9.17$ \\
        & Math & $22.27$ & $37.17$ & $50.00$ & $33.56$ & $29.90$ & $19.29$ & $25.21$ & $17.12$ & $16.70$ \\
        \midrule
        OpenAI-o3-mini.high.code & Biology & $41.16$ & $72.11$ & $72.16$ & $75.00$ & $60.00$ & $34.97$ & $39.59$ & $23.50$ & $36.00$ \\
        & Chemistry & $17.46$ & $23.13$ & $30.61$ & $18.37$ & $20.41$ & $16.33$ & $17.96$ & $14.29$ & $16.73$ \\
        & Physics & $18.46$ & $32.04$ & $42.86$ & $26.57$ & $25.00$ & $15.75$ & $20.17$ & $13.29$ & $13.33$ \\
        & Math & $29.35$ & $47.79$ & $57.29$ & $45.21$ & $42.27$ & $25.66$ & $31.04$ & $22.88$ & $24.54$ \\
        \midrule
        Gemini-2.5-flash.preview.0520 & Biology & $47.96$ & $74.83$ & $75.26$ & $77.50$ & $60.00$ & $42.59$ & $46.60$ & $34.50$ & $36.00$ \\
        & Chemistry &$19.95$ & $31.29$  & $34.69$ & $28.57$ & $30.61$ & $17.69$ & $19.18$ & $17.55$ & $16.33$ \\
        & Physics & $18.41$ &  $31.74$ &  $42.02$ & $27.97$ & $22.22$ & $15.75$ &  $21.18$ & $12.17$ & $13.89$ \\
        & Math & $21.44$ & $40.41$ & $46.88$ & $41.78$ & $31.96$ & $17.64$ & $19.58$ & $17.81$ & $15.46$ \\
        \midrule
        GPT4o-1120 & Biology & $36.17$ & $61.90$ & $63.92$ & $65.00$ & $30.00$ & $31.02$ & $33.20$ & $30.00$ & $14.00$ \\
        & Chemistry & $14.51$ & $23.81$ & $30.61$ & $24.49$ & $16.33$ & $12.65$ & $19.59$ & $6.12$ & $12.24$ \\
        & Physics & $10.63$ & $19.76$ & $31.09$ & $16.78$ & $6.94$ & $8.80$ & $12.94$ & $7.55$ & $4.44$ \\
        & Math & $9.83$ & $24.78$ & $35.42$ & $23.29$ & $16.49$ & $6.84$ & $9.58$ & $7.53$ & $3.09$ \\
        \midrule
        GPT4o-0513 & Biology & $36.05$ & $59.86$ & $61.86$ & $62.50$ & $30.00$ & $31.29$ & $33.40$ & $30.00$ & $16.00$ \\
        & Chemistry & $15.53$ & $22.45$ & $30.61$ & $22.45$ & $14.29$ & $14.15$ & $23.27$ & $11.02$ & $8.16$ \\
        & Physics & $9.88$ & $20.06$ & $33.61$ & $16.08$ & $5.56$ & $7.84$ & $13.11$ & $6.43$ & $1.94$ \\
        & Math & $9.78$ & $23.01$ & $33.33$ & $19.86$ & $17.53$ & $7.14$ & $10.00$ & $7.67$ & $3.51$ \\
        \midrule
        GPT4o-0806 & Biology & $34.81$ & $57.82$ & $63.92$ & $55.00$ & $10.00$ & $30.20$ & $34.02$ & $25.00$ & $14.00$ \\
        & Chemistry & $12.59$ & $19.05$ & $32.65$ & $8.16$ & $16.33$ & $11.29$ & $19.59$ & $7.76$ & $6.53$ \\
        & Physics & $8.23$ & $19.76$ & $33.61$ & $14.69$ & $6.94$ & $5.93$ & $9.08$ & $4.76$ & $3.06$ \\
        & Math & $8.31$ & $20.65$ & $28.13$ & $19.86$ & $14.43$ & $5.84$ & $7.71$ & $6.30$ & $3.30$ \\
        \bottomrule
    \end{tabular}
    }
\end{table}

%% file: sections/discussion.tex
\section{Discussion}

\subsection{Trained problems cannot robustly generalize to other problems with the same solution approach}

To further analyze why the model fails to correctly solve problems after random parameter initialization, we conducted a meticulous manual verification of its incorrect answers. For each subject, we selected one incorrect answer generated by either the thinking model (OpenAI-o3-high.code) or the instruct model (GPT4o-1120) for detailed human scrutiny. Specifically, we randomly sampled 50 problems from each subject (35 in chemistry for GPT4o-1120) and examined all of their corresponding incorrect responses.

We categorized the identified errors based on the following criteria: Logical Errors encompassed issues where the calculation method was incorrect, an erroneous formula was applied, or the reasoning process contained fundamental flaws, indicating the model's failure to grasp the problem or apply appropriate solution strategies. Calculation Errors, on the other hand, included unit confusion, prevalent answer precision problems, incorrect intermediate numerical calculations, or minor computational missteps despite the overall method being correct, suggesting the model struggled with the execution of valid solution steps.



\begin{figure}[htbp]
    \centering
    \includegraphics[width=1.0\textwidth]{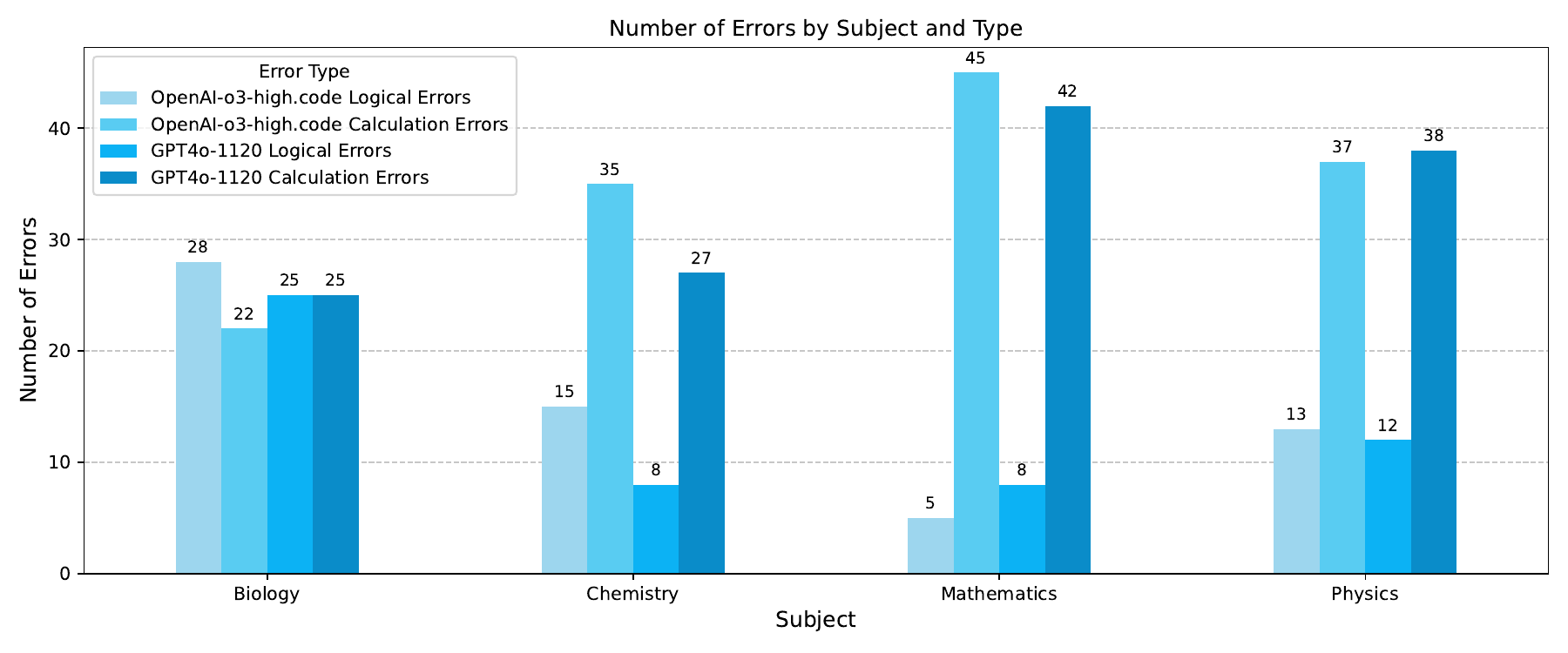} 
    \caption{Distribution of error types for different models across various subjects.}
    \label{fig:error_statistics}
\end{figure}

The statistic results are shown in Figure \ref{fig:error_statistics}. A consistent trend in the distribution of error types was observed across both models. In all disciplines other than Biology, calculation errors were dominant, constituting at least two-thirds of all errors. This indicates that the models were likely trained on larger corpora for these subjects, resulting in superior generalization. Therefore, the main bottleneck appears to be the models' computational capacity. Conversely, for subjects such as biology where corpus is relatively scarce, weaker generalization capabilities likely cause a higher incidence of logical errors, leading them to occur at a frequency nearly equivalent to that of calculation errors. 

In summary, we suggest that a model's error types on problems with randomized parameters can, to some extent, reflect its generalization ability in the corresponding discipline. While better generalization leads to fewer logical reasoning errors, challenges with calculation and instruction-following persist. Furthermore, this generalization capability is likely correlated with the richness of the relevant training data.

\subsection{The use of Code Interpreter (CI) arithmetic is necessary to maintain numerical stability.}

The experimental data unequivocally highlights the profound impact of integrating a Code Interpreter (CI) on the model's computational robustness and overall performance. The most striking insight gleaned from these results is that the presence of CI leads to a substantial increase in the model's overall average score. We utilize Gemini-2.5-pro.preview to perform the control experiment. Specifically, Gemini-2.5-pro.preview.0506.google.ci achieved an average score of 40.19, markedly higher than the 30.20 recorded by Gemini-2.5-pro.preview.0506 without CI. This significant improvement across the board underscores CI's ability to enhance problem-solving capabilities, primarily by ensuring more precise arithmetic computations.

Furthermore, the analysis reveals that the positive influence of CI extends across all difficulty levels and both parameter initialization methods. Whether the parameters were initialized in a standard manner ('Initial') or randomly ('Random'), the CI-enabled model consistently demonstrated superior performance. For instance, under initial parameter settings, Gemini-2.5-pro.preview.0506.google.ci outperformed its non-CI counterpart across easy (55.68 vs. 53.19), medium (48.68 vs. 46.30), and hard (42.54 vs. 34.65) problems. A similar, if not more pronounced, trend was observed under random parameter initialization. This consistent uplift across varied conditions emphasizes CI's role in bolstering the model's inherent numerical stability and generalization capabilities.


Crucially, the impact of CI is particularly pronounced when tackling difficult problems. The most substantial performance gains were observed in the 'Hard' category, under both 'Initial' and 'Random' parameter settings. For instance, with initial parameters, the CI model's score of 42.54 on hard problems was a significant leap from the non-CI model's 34.65, representing an approximate 22.79\% increase. Similarly, under random initialization, the CI model achieved 32.63 on hard problems, considerably higher than the 22.89 without CI, an improvement of roughly 42.53\%. This suggests that for complex numerical operations, where internal model computations might be prone to precision errors or instability, the external precision provided by the Code Interpreter becomes absolutely critical, allowing the model to maintain accuracy and achieve higher success rates on challenging tasks.

%% file: sections/conclusion.tex
\section{Conclusion}
To conduct comprehensive and truthful assesment of LLMs reasoning capabilities without data contamination, a comprehensive and challenging dynamic scientific benchmark holds significance. Therefore, we proposed \textbf{SciDA}, a multi-disciplinary benchmark that consists exclusively of over 1k Olympic-level numerical computation problems, allowing randomized numerical initializations for each inference round to disrupt memorization patterns and void reliance on fixed numerical patterns. Thus we ensures that the cognitive reasoning and problem-solving capabilities of LLMs are accurately assessed without bias. 

We conduct a series of experiments with both closed-source and open-source top-performing LLMs, and it is observed that the performance of LLMs drop significantly under random numerical initialization. Such result strongly indicates the widespread presence of data contamination within LLMs. The superior performance under initial condition suggests that LLMs may have encounterd similar problems instances or memorized numerical patterns during training, leading to inflated score. Conversely, when the parameters of problems are dynamically initialized, such problems would not occur, demanding true generalization from the models. 

To conclude, we have proposed a new paradigm of scientific benchmark allowing dynamical initializations to mitigate data contamination. Our work features broad discipline coverage and expert-annotated high-quality problems, which facilitates truthful assessment of LLMs scientific reasoning capabilities. As LLMs have achieved remarkable performance in tasks across various branches of the natural sciences, our work revealed the long-exsiting over-estimation of their capabilities. Moreover, we believe that our work would undoubtedly play a role in narrowing the gap between talented human scientists and LLMs and such advancement towards Artificial General Intelligence (AGI) could facilitate the possibility of LLMs to advance the frontier of human knowledge.

%% file: sections/future_work.tex
\section{Future Work}
We are actively working to expand the scale and disciplinary coverage of SciDA. Our goal is to extend beyond the common STEM subjects—such as mathematics, physics, chemistry, and biology—to include a wider variety of disciplines. This expansion will enable a more thorough evaluation of LLMs' performance on diverse data, thereby establishing SciDA as a valuable and comprehensive benchmark dataset for the LLM community.

%% file: sections/appendix.tex
\section{Contributor \& Acknowledgement}

Junting Zhou\textsuperscript{2,4,*},
Tingjia Miao\textsuperscript{3,*},  
Yiyan Liao\textsuperscript{2}, 
Qichao Wang\textsuperscript{5},  
Zhoufutu Wen\textsuperscript{1,4},  
Yanqin Wang\textsuperscript{2}, 
Yunjie Huang\textsuperscript{3},  
Ge Yan\textsuperscript{3},  
Leqi Wang\textsuperscript{3},  
Yucheng Xia\textsuperscript{3},  
Hongwan Gao\textsuperscript{1},  
Yuansong Zeng\textsuperscript{1},  
Renjie Zheng\textsuperscript{1},
Chen Dun\textsuperscript{1},
Yitao Liang\textsuperscript{1,\dag},
Tong Yang\textsuperscript{2,\dag},  
Wenhao Huang\textsuperscript{1,4,\dag},  
Ge Zhang\textsuperscript{1,4,\dag}  

\vspace{1em}

\textsuperscript{1}ByteDance Seed  
\textsuperscript{2}Peking University  
\textsuperscript{3}Shanghai Jiao Tong University  
\textsuperscript{4}M-A-P
\textsuperscript{5}Jilin University  

\vspace{1em}

\textsuperscript{*}Equal Contribution  
\textsuperscript{\dag}Corresponding authors

\section{Data Source}
Our data covers various disciplines and includes both publicly available and privately held or original Olympic-level problems.

We have meticulously sourced problems that meet our requirements from regional and international Olympiad competition problems, Olympiad workbook and guides, and professional college textbooks. The publicly available sources includes:

\begin{enumerate}
    \item \textbf{Mathematics}: International Mathematical Olympiad (IMO), Chinese Mathematical Olympiad (CMO), Problems in Mathematical Analysis by B. P. Demidovich, Euler Math, etc.
    \item \textbf{Physics}: International Physics Olympiad (IPhO), Chinese Mathematical Olympiad (CPhO), International Physics Olympiad Training and Selection by Yongling Zheng, Collection of Physics Challenges by Yousheng Shu et al., A Grand Dictionary of Plysics Prolens and Solutons by Yongde Zhang et al., New Concept Physics Tutorial by Kaihua Zhao et al., Mechanics by Yousheng Shu et al., etc., Modern Quantum Mechenics by Sakurai Jun, Quantum Mechenics Solution Manual by David J. Griffiths, Electrodynamics Solution Manual by David J. Griffiths, etc.
    \item \textbf{Chemistry}: International Chemistry Olympiad (IChO), Chinese Chemistry Olympiad (CChO), Physical Chemistry by Peter Atkins, etc.
    \item \textbf{Biology}: International Biology Olympiad (IBO), Chinese National Biology Olympiad (CNBO), etc.
\end{enumerate}

High-quality privately held or original problems constitute another pillar of our testing benchmark. These problems are contributed by trusted Olympic competition medalists, coaches, and university professors, accounting for over 20\% of the total data volume, while the proportion is higher in chemistry and biology.

\section{Elbow Point Analysis for Optimal Random Sampling Number}
\label{appendix:Elbow_Analysis}
In our main analysis, we performed 5 random parameter initializations for each problem. Here, we present the elbow point analysis conducted to determine the optimal number of random initializations, denoted as n.

Specifically, we utilized two models: GPT-4o-1120 as the instruction model and OpenAI-o3-mini.high.code as the thinking model. For each model, we conducted 10 independent random parameter initializations and performed inference accordingly. From the 10 resulting sets of outputs for each model, we created subgroups of size n, where n ranged from 2 to 10. For each value of n, we repeatedly sampled n sets from the 10 sets and calculated the mean of these samples. We then computed the variance of this distribution of means. By plotting the relationship between n and the variance of the means, we identified the "elbow point", which represents the optimal random sampling number.

\begin{figure}[h!tbp]
    \centering
    \begin{subfigure}[b]{0.48\textwidth}
        \includegraphics[width=\textwidth]{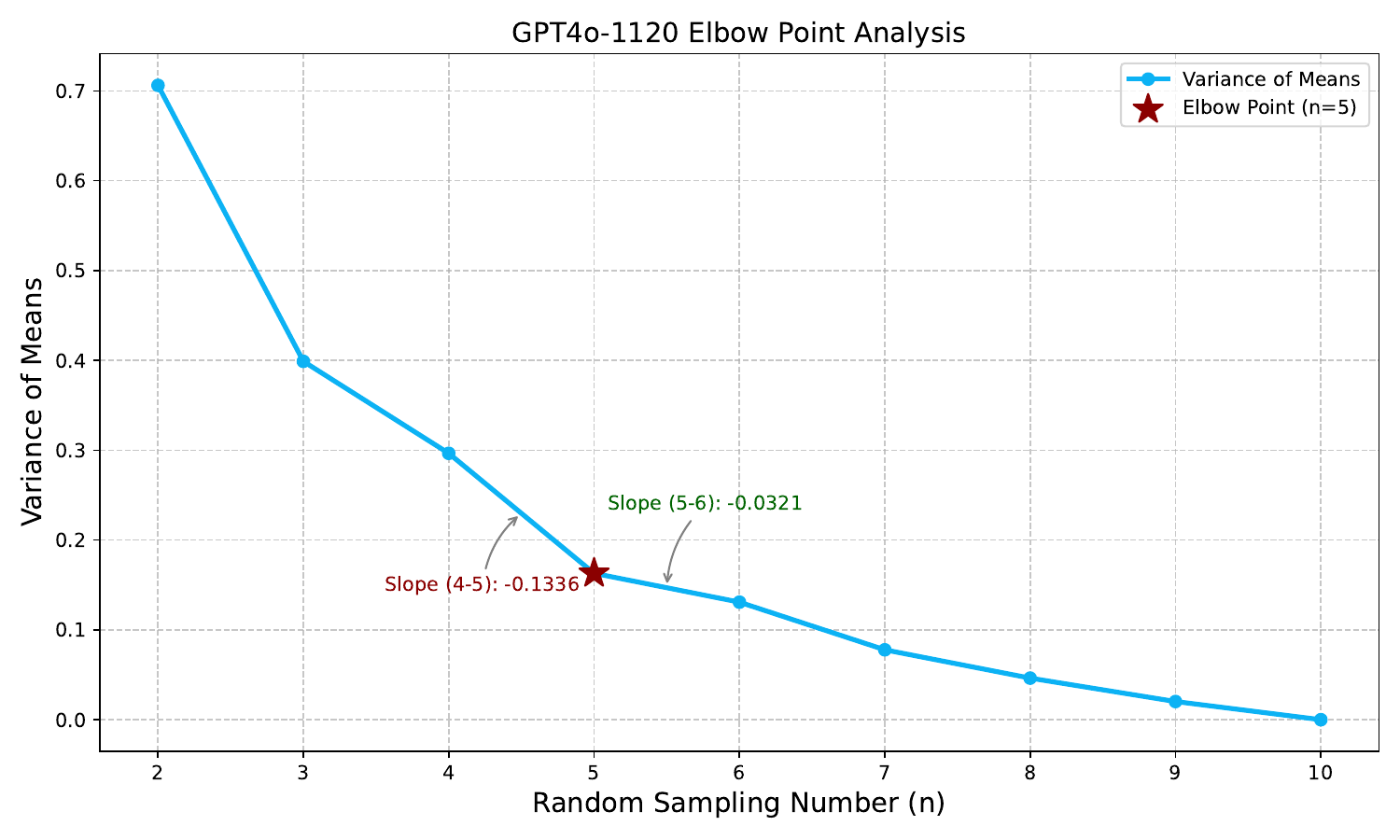}
        \caption{Elbow point analysis for GPT4o-1120 model showing the relationship between the variance of means and the random sampling number (n).}
        \label{fig:GPT4o-1120_elbow}
    \end{subfigure}
    \hfill
    \begin{subfigure}[b]{0.48\textwidth}
        \includegraphics[width=\textwidth]{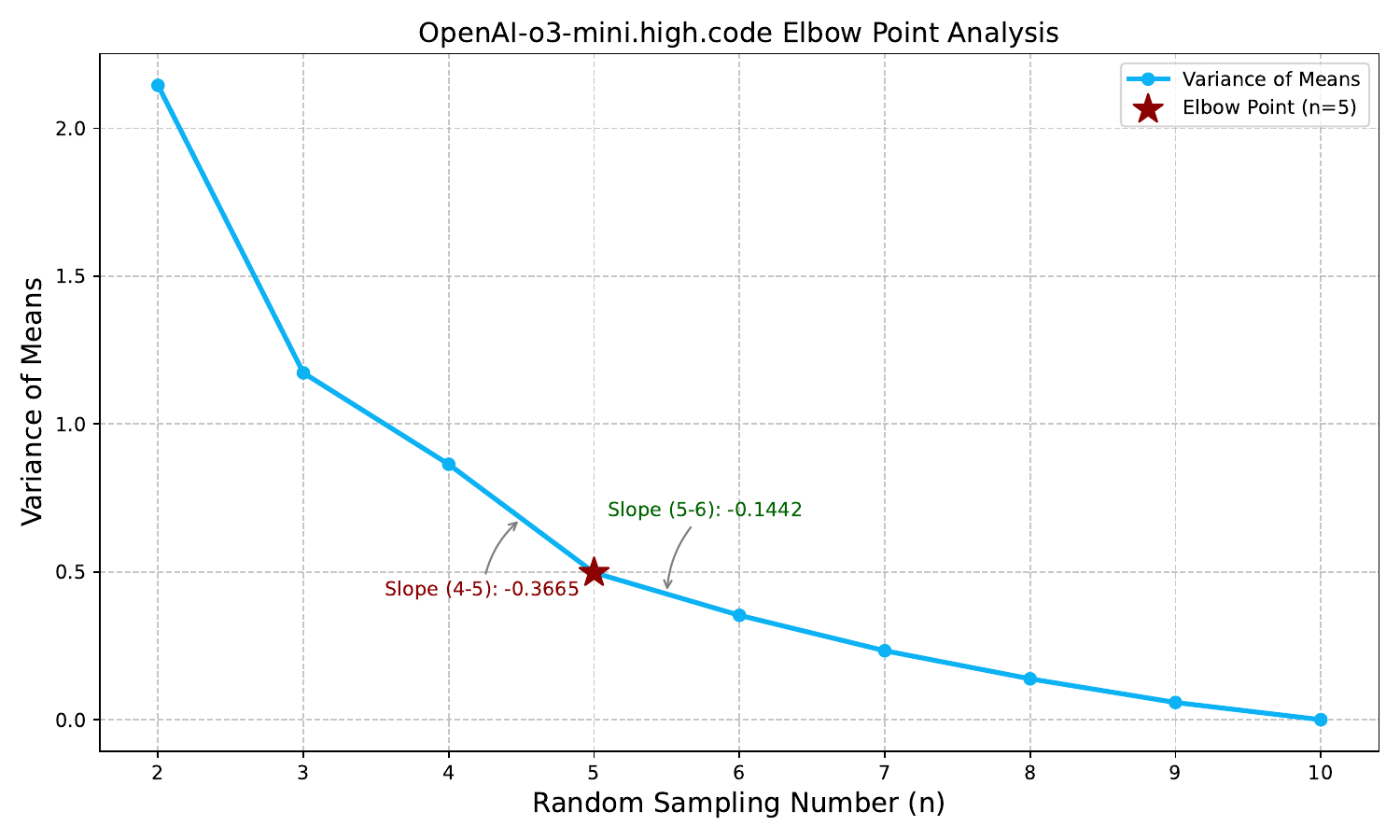}
        \caption{Elbow point analysis for OpenAI-o3-mini.high.code model showing the relationship between the variance of means and the random sampling number (n).}
        \label{fig:OpenAI-o3-mini.high.code_elbow}
    \end{subfigure}
    \caption{Elbow point analysis for optimal random sampling number.}
    \label{fig:combined_elbow}
\end{figure}

The results are displayed in Figure \ref{fig:combined_elbow}. As can be seen, n=5 is the elbow point for the inference results of both models. This value represents the most robust and cost-effective choice. Increasing the number of random initializations to 6 or more yields diminishing returns, as the marginal benefit in performance does not justify the additional time and computational costs. Therefore, we conclude that the optimal random sampling number is 5, as it strikes the best balance between resource consumption for inference and the accuracy of the model evaluation.

%% file: paper.bbl
\begin{thebibliography}{23}
\providecommand{\natexlab}[1]{#1}
\providecommand{\url}[1]{\texttt{#1}}
\expandafter\ifx\csname urlstyle\endcsname\relax
  \providecommand{\doi}[1]{doi: #1}\else
  \providecommand{\doi}{doi: \begingroup \urlstyle{rm}\Url}\fi

\bibitem[Cobbe et~al.(2021)Cobbe, Kosaraju, Bavarian, Chen, Jun, Kaiser, Plappert, Tworek, Hilton, Nakano, Hesse, and Schulman]{cobbe2021verifier}
Karl Cobbe, Vineet Kosaraju, Mohammad Bavarian, Mark Chen, Heewoo Jun, Lukasz Kaiser, Matthias Plappert, Jerry Tworek, Jacob Hilton, Reiichiro Nakano, Christopher Hesse, and John Schulman.
\newblock Training verifiers to solve math word problems, 2021.
\newblock URL \url{https://arxiv.org/abs/2110.14168}.

\bibitem[Deng et~al.(2023)Deng, Zhao, Tang, Gerstein, and Cohan]{deng2023conta}
Chunyuan Deng, Yilun Zhao, Xiangru Tang, Mark Gerstein, and Arman Cohan.
\newblock Investigating data contamination in modern benchmarks for large language models.
\newblock \emph{arXiv preprint arXiv:2311.09783}, 2023.

\bibitem[Dong et~al.(2024)Dong, Jiang, Liu, Jin, Gu, Yang, and Li]{dong2024generalization}
Yihong Dong, Xue Jiang, Huanyu Liu, Zhi Jin, Bin Gu, Mengfei Yang, and Ge~Li.
\newblock Generalization or memorization: Data contamination and trustworthy evaluation for large language models.
\newblock \emph{arXiv preprint arXiv:2402.15938}, 2024.

\bibitem[Gao et~al.(2024)Gao, Song, Yang, Cai, Miao, Dong, Li, Ma, Chen, Xu, Tang, Wang, Zan, Quan, Zhang, Sha, Zhang, Ren, Liu, and Chang]{gao2024omnimath}
Bofei Gao, Feifan Song, Zhe Yang, Zefan Cai, Yibo Miao, Qingxiu Dong, Lei Li, Chenghao Ma, Liang Chen, Runxin Xu, Zhengyang Tang, Benyou Wang, Daoguang Zan, Shanghaoran Quan, Ge~Zhang, Lei Sha, Yichang Zhang, Xuancheng Ren, Tianyu Liu, and Baobao Chang.
\newblock Omni-math: A universal olympiad level mathematic benchmark for large language models, 2024.
\newblock URL \url{https://arxiv.org/abs/2410.07985}.

\bibitem[Golchin and Surdeanu(2023)]{golchin2023time}
Shahriar Golchin and Mihai Surdeanu.
\newblock Time travel in llms: Tracing data contamination in large language models.
\newblock \emph{arXiv preprint arXiv:2308.08493}, 2023.

\bibitem[He et~al.(2024)He, Luo, Bai, Hu, Thai, Shen, Hu, Han, Huang, Zhang, et~al.]{he2024olympiadbench}
Chaoqun He, Renjie Luo, Yuzhuo Bai, Shengding Hu, Zhen~Leng Thai, Junhao Shen, Jinyi Hu, Xu~Han, Yujie Huang, Yuxiang Zhang, et~al.
\newblock Olympiadbench: A challenging benchmark for promoting agi with olympiad-level bilingual multimodal scientific problems.
\newblock \emph{arXiv preprint arXiv:2402.14008}, 2024.

\bibitem[Hendrycks et~al.(2021{\natexlab{a}})Hendrycks, Burns, Basart, Zou, Mazeika, Song, and Steinhardt]{hendrycks2021mmlu}
Dan Hendrycks, Collin Burns, Steven Basart, Andy Zou, Mantas Mazeika, Dawn Song, and Jacob Steinhardt.
\newblock Measuring massive multitask language understanding, 2021{\natexlab{a}}.
\newblock URL \url{https://arxiv.org/abs/2009.03300}.

\bibitem[Hendrycks et~al.(2021{\natexlab{b}})Hendrycks, Burns, Kadavath, Arora, Basart, Tang, Song, and Steinhardt]{hendrycks2021math}
Dan Hendrycks, Collin Burns, Saurav Kadavath, Akul Arora, Steven Basart, Eric Tang, Dawn Song, and Jacob Steinhardt.
\newblock Measuring mathematical problem solving with the math dataset, 2021{\natexlab{b}}.
\newblock URL \url{https://arxiv.org/abs/2103.03874}.

\bibitem[Huang et~al.(2025{\natexlab{a}})Huang, Guo, Li, Ji, Ge, Li, Guo, Cai, Yuan, Wang, Wu, Yin, Tang, Huang, Jin, Chen, Zhang, and Wang]{huang2025mathperturbbenchmarkingllmsmath}
Kaixuan Huang, Jiacheng Guo, Zihao Li, Xiang Ji, Jiawei Ge, Wenzhe Li, Yingqing Guo, Tianle Cai, Hui Yuan, Runzhe Wang, Yue Wu, Ming Yin, Shange Tang, Yangsibo Huang, Chi Jin, Xinyun Chen, Chiyuan Zhang, and Mengdi Wang.
\newblock Math-perturb: Benchmarking llms' math reasoning abilities against hard perturbations, 2025{\natexlab{a}}.
\newblock URL \url{https://arxiv.org/abs/2502.06453}.

\bibitem[Huang et~al.(2025{\natexlab{b}})Huang, Wang, Xia, Li, Zou, Xu, Fan, Ye, Chern, Ye, Zhang, Yang, Wu, Wang, Sun, Xiao, Li, Zhou, Chern, Qin, Ma, Su, Liu, Zheng, Zhang, Lin, Qiao, and Liu]{huang2025olympicarena}
Zhen Huang, Zengzhi Wang, Shijie Xia, Xuefeng Li, Haoyang Zou, Ruijie Xu, Run-Ze Fan, Lyumanshan Ye, Ethan Chern, Yixin Ye, Yikai Zhang, Yuqing Yang, Ting Wu, Binjie Wang, Shichao Sun, Yang Xiao, Yiyuan Li, Fan Zhou, Steffi Chern, Yiwei Qin, Yan Ma, Jiadi Su, Yixiu Liu, Yuxiang Zheng, Shaoting Zhang, Dahua Lin, Yu~Qiao, and Pengfei Liu.
\newblock Olympicarena: Benchmarking multi-discipline cognitive reasoning for superintelligent ai, 2025{\natexlab{b}}.
\newblock URL \url{https://arxiv.org/abs/2406.12753}.

\bibitem[Jain et~al.(2025)Jain, Han, Gu, Li, Yan, Zhang, Wang, Solar-Lezama, Sen, and Stoica]{jain2025livecodebench}
Naman Jain, King Han, Alex Gu, Wen-Ding Li, Fanjia Yan, Tianjun Zhang, Sida Wang, Armando Solar-Lezama, Koushik Sen, and Ion Stoica.
\newblock Livecodebench: Holistic and contamination free evaluation of large language models for code.
\newblock In \emph{The Thirteenth International Conference on Learning Representations}, 2025.
\newblock URL \url{https://openreview.net/forum?id=chfJJYC3iL}.

\bibitem[Liu et~al.(2024)Liu, Zheng, Qiao, Duan, Fei, Zhou, Zhang, Zhang, Lin, and Chen]{liu2024mathbench}
Hongwei Liu, Zilong Zheng, Yuxuan Qiao, Haodong Duan, Zhiwei Fei, Fengzhe Zhou, Wenwei Zhang, Songyang Zhang, Dahua Lin, and Kai Chen.
\newblock Mathbench: Evaluating the theory and application proficiency of llms with a hierarchical mathematics benchmark.
\newblock \emph{arXiv preprint arXiv:2405.12209}, 2024.

\bibitem[Qian et~al.(2024)Qian, Wan, Tang, Wang, Zhang, Chen, and Yu]{qian2024varbench}
Kun Qian, Shunji Wan, Claudia Tang, Youzhi Wang, Xuanming Zhang, Maximillian Chen, and Zhou Yu.
\newblock Varbench: Robust language model benchmarking through dynamic variable perturbation.
\newblock \emph{arXiv preprint arXiv:2406.17681}, 2024.

\bibitem[Qiu et~al.(2025)Qiu, Guo, Song, Sun, Cai, Wei, Luo, Yin, Zhang, Hu, Wang, Tang, Chang, Liu, Zhou, Zhang, Zhang, Liu, Li, Zhang, Jing, Yin, Ren, Fu, Ji, Wang, Tian, Lv, Man, Li, Tao, Sun, Liang, Mu, Li, Zhang, Zhang, Li, Xia, Lin, Shen, Chen, Xiong, Wang, Wang, Ni, Zhang, Cui, Shao, Cao, xing Luo, Yang, Zhang, and Zhu]{qiu2025phybench}
Shi Qiu, Shaoyang Guo, Zhuo-Yang Song, Yunbo Sun, Zeyu Cai, Jiashen Wei, Tianyu Luo, Yixuan Yin, Haoxu Zhang, Yi~Hu, Chenyang Wang, Chencheng Tang, Haoling Chang, Qi~Liu, Ziheng Zhou, Tianyu Zhang, Jingtian Zhang, Zhangyi Liu, Minghao Li, Yuku Zhang, Boxuan Jing, Xianqi Yin, Yutong Ren, Zizhuo Fu, Jiaming Ji, Weike Wang, Xudong Tian, Anqi Lv, Laifu Man, Jianxiang Li, Feiyu Tao, Qihua Sun, Zhou Liang, Yushu Mu, Zhongxuan Li, Jing-Jun Zhang, Shutao Zhang, Xiaotian Li, Xingqi Xia, Jiawei Lin, Zheyu Shen, Jiahang Chen, Qiuhao Xiong, Binran Wang, Fengyuan Wang, Ziyang Ni, Bohan Zhang, Fan Cui, Changkun Shao, Qing-Hong Cao, Ming xing Luo, Yaodong Yang, Muhan Zhang, and Hua~Xing Zhu.
\newblock Phybench: Holistic evaluation of physical perception and reasoning in large language models, 2025.
\newblock URL \url{https://arxiv.org/abs/2504.16074}.

\bibitem[Rein et~al.(2024)Rein, Hou, Stickland, Petty, Pang, Dirani, Michael, and Bowman]{rein2024gpqa}
David Rein, Betty~Li Hou, Asa~Cooper Stickland, Jackson Petty, Richard~Yuanzhe Pang, Julien Dirani, Julian Michael, and Samuel~R. Bowman.
\newblock {GPQA}: A graduate-level google-proof q\&a benchmark.
\newblock In \emph{First Conference on Language Modeling}, 2024.
\newblock URL \url{https://openreview.net/forum?id=Ti67584b98}.

\bibitem[Shi et~al.(2025)Shi, Yang, Liu, Bu, Chen, Zhou, Ma, Wen, Wang, He, Song, Zhu, Li, Wang, Zhang, Yuan, Yao, Yang, Wang, Fang, Yuan, He, Tang, Tan, Zhou, Zhang, Li, Huang, and Zhang]{shi2025korgymdynamicgameplatform}
Jiajun Shi, Jian Yang, Jiaheng Liu, Xingyuan Bu, Jiangjie Chen, Junting Zhou, Kaijing Ma, Zhoufutu Wen, Bingli Wang, Yancheng He, Liang Song, Hualei Zhu, Shilong Li, Xingjian Wang, Wei Zhang, Ruibin Yuan, Yifan Yao, Wenjun Yang, Yunli Wang, Siyuan Fang, Siyu Yuan, Qianyu He, Xiangru Tang, Yingshui Tan, Wangchunshu Zhou, Zhaoxiang Zhang, Zhoujun Li, Wenhao Huang, and Ge~Zhang.
\newblock Korgym: A dynamic game platform for llm reasoning evaluation, 2025.
\newblock URL \url{https://arxiv.org/abs/2505.14552}.

\bibitem[Sun et~al.(2025)Sun, Min, Chen, Zhao, Fang, Liu, Wang, and Wen]{sun2025olymmath}
Haoxiang Sun, Yingqian Min, Zhipeng Chen, Wayne~Xin Zhao, Lei Fang, Zheng Liu, Zhongyuan Wang, and Ji-Rong Wen.
\newblock Challenging the boundaries of reasoning: An olympiad-level math benchmark for large language models, 2025.
\newblock URL \url{https://arxiv.org/abs/2503.21380}.

\bibitem[Sun et~al.(2024)Sun, Han, Zhao, Ma, Shen, Chen, Chen, and Yu]{sun2024scieval}
Liangtai Sun, Yang Han, Zihan Zhao, Da~Ma, Zhennan Shen, Baocai Chen, Lu~Chen, and Kai Yu.
\newblock Scieval: A multi-level large language model evaluation benchmark for scientific research.
\newblock In \emph{Proceedings of the AAAI Conference on Artificial Intelligence}, volume~38, pages 19053--19061, 2024.

\bibitem[Tang et~al.(2024)Tang, Zhang, Wang, and Wei]{tang2024mathscale}
Zhengyang Tang, Xingxing Zhang, Benyou Wang, and Furu Wei.
\newblock Mathscale: Scaling instruction tuning for mathematical reasoning.
\newblock \emph{arXiv preprint arXiv:2403.02884}, 2024.

\bibitem[Team et~al.(2025)Team, Du, Yao, Ma, Wang, Zheng, Zhu, Liu, Liang, Jin, Wei, Zheng, Deng, Gavin, Jia, Jiang, Liao, Li, Li, Li, Li, Li, Ma, Ni, Que, Wang, Wen, Wu, Hsing, Xu, Yang, Wang, Zhou, Bai, Bu, Cai, Chen, Chen, Cheng, Cheng, Ding, Huang, Huang, Li, Li, Li, Liang, Lin, Lin, Ma, Pang, Peng, Peng, Qi, Qiu, Qu, Quan, Tan, Wang, Wang, Wang, Wang, Wang, Xu, Yang, Yuan, Yue, Zhan, Zhang, Zhang, Zhang, Zhang, Zhang, Zhao, Zheng, Zhong, Gao, Li, Liu, Liu, Liu, Ni, Peng, Qin, Su, Wang, Wang, Yang, Yang, Cao, Yue, Zhang, Zhou, Liu, Lin, Huang, and Zhang]{pteam2025supergpqascalingllmevaluation}
P~Team, Xinrun Du, Yifan Yao, Kaijing Ma, Bingli Wang, Tianyu Zheng, King Zhu, Minghao Liu, Yiming Liang, Xiaolong Jin, Zhenlin Wei, Chujie Zheng, Kaixin Deng, Shawn Gavin, Shian Jia, Sichao Jiang, Yiyan Liao, Rui Li, Qinrui Li, Sirun Li, Yizhi Li, Yunwen Li, David Ma, Yuansheng Ni, Haoran Que, Qiyao Wang, Zhoufutu Wen, Siwei Wu, Tyshawn Hsing, Ming Xu, Zhenzhu Yang, Zekun~Moore Wang, Junting Zhou, Yuelin Bai, Xingyuan Bu, Chenglin Cai, Liang Chen, Yifan Chen, Chengtuo Cheng, Tianhao Cheng, Keyi Ding, Siming Huang, Yun Huang, Yaoru Li, Yizhe Li, Zhaoqun Li, Tianhao Liang, Chengdong Lin, Hongquan Lin, Yinghao Ma, Tianyang Pang, Zhongyuan Peng, Zifan Peng, Qige Qi, Shi Qiu, Xingwei Qu, Shanghaoran Quan, Yizhou Tan, Zili Wang, Chenqing Wang, Hao Wang, Yiya Wang, Yubo Wang, Jiajun Xu, Kexin Yang, Ruibin Yuan, Yuanhao Yue, Tianyang Zhan, Chun Zhang, Jinyang Zhang, Xiyue Zhang, Xingjian Zhang, Yue Zhang, Yongchi Zhao, Xiangyu Zheng, Chenghua Zhong, Yang Gao, Zhoujun Li, Dayiheng Liu, Qian Liu, Tianyu Liu, Shiwen
  Ni, Junran Peng, Yujia Qin, Wenbo Su, Guoyin Wang, Shi Wang, Jian Yang, Min Yang, Meng Cao, Xiang Yue, Zhaoxiang Zhang, Wangchunshu Zhou, Jiaheng Liu, Qunshu Lin, Wenhao Huang, and Ge~Zhang.
\newblock Supergpqa: Scaling llm evaluation across 285 graduate disciplines, 2025.
\newblock URL \url{https://arxiv.org/abs/2502.14739}.

\bibitem[Wang et~al.(2023)Wang, Hu, Lu, Zhu, Zhang, Subramaniam, Loomba, Zhang, Sun, and Wang]{wang2023scibench}
Xiaoxuan Wang, Ziniu Hu, Pan Lu, Yanqiao Zhu, Jieyu Zhang, Satyen Subramaniam, Arjun~R Loomba, Shichang Zhang, Yizhou Sun, and Wei Wang.
\newblock Scibench: Evaluating college-level scientific problem-solving abilities of large language models.
\newblock \emph{arXiv preprint arXiv:2307.10635}, 2023.

\bibitem[Wang et~al.(2024)Wang, Ma, Zhang, Ni, Chandra, Guo, Ren, Arulraj, He, Jiang, Li, Ku, Wang, Zhuang, Fan, Yue, and Chen]{wang2024mmlupro}
Yubo Wang, Xueguang Ma, Ge~Zhang, Yuansheng Ni, Abhranil Chandra, Shiguang Guo, Weiming Ren, Aaran Arulraj, Xuan He, Ziyan Jiang, Tianle Li, Max Ku, Kai Wang, Alex Zhuang, Rongqi Fan, Xiang Yue, and Wenhu Chen.
\newblock {MMLU}-pro: A more robust and challenging multi-task language understanding benchmark.
\newblock In \emph{The Thirty-eight Conference on Neural Information Processing Systems Datasets and Benchmarks Track}, 2024.
\newblock URL \url{https://openreview.net/forum?id=y10DM6R2r3}.

\bibitem[Zhong et~al.(2023)Zhong, Cui, Guo, Liang, Lu, Wang, Saied, Chen, and Duan]{zhong2023agieval}
Wanjun Zhong, Ruixiang Cui, Yiduo Guo, Yaobo Liang, Shuai Lu, Yanlin Wang, Amin Saied, Weizhu Chen, and Nan Duan.
\newblock Agieval: A human-centric benchmark for evaluating foundation models.
\newblock \emph{arXiv preprint arXiv:2304.06364}, 2023.

\end{thebibliography}
